\pdfoutput=1

\documentclass[11pt]{article}

\usepackage{emnlp2021}

\usepackage{times}
\usepackage{latexsym}

\usepackage[T1]{fontenc}

\usepackage[utf8]{inputenc}

\usepackage{microtype}
\usepackage{amsfonts}
\usepackage{amsmath}

\usepackage{booktabs}
\usepackage{tabularx}
\newcolumntype{Y}{>{\centering\arraybackslash}X}
\usepackage{multirow}

\usepackage{cleveref}
\crefname{section}{\S}{\S\S}
\Crefname{section}{\S}{\S\S}
\crefname{table}{Tab.}{}
\crefname{figure}{Fig.}{}
\crefname{algorithm}{Alg.}{}
\crefname{appendix}{App.}{}
\crefname{lemma}{Lemma}{}
\Crefname{theorem}{Theorem}{}
\crefname{prop}{Proposition}{}
\crefname{cor}{Corollary}{}
\crefname{align}{}{}
\crefname{equation}{}{}

\usepackage{tikz}
\usepackage{pgfplots}
\usetikzlibrary{shapes.geometric, arrows, positioning}
\tikzstyle{arrow} = [thick,->,>=stealth]

\usepackage{xcolor}

\title{Vision Matters When It Should:\\ Sanity Checking Multimodal Machine Translation Models}

\author{Jiaoda Li$^{1}$ \qquad Duygu Ataman$^{2}\thanks{\hspace{2mm}Work done while at the University of Z{\"u}rich.}$ \qquad Rico Sennrich$^{3,4}$ \bigskip\\
$^1$ETH Z{\"u}rich \\ $^2$New York University \\ $^3$University of Z{\"u}rich \\ $^4$University of Edinburgh \\
{\tt jiaoda.li@inf.ethz.ch} \quad {\tt ataman@nyu.edu} \quad {\tt sennrich@cl.uzh.ch} \quad 
}

\begin{document}
\maketitle
\begin{abstract}
Multimodal machine translation (MMT) systems have been shown to outperform their text-only neural machine translation (NMT) counterparts when visual context is available. However, recent studies have also shown that the performance of MMT models is only marginally impacted when the associated image is replaced with an unrelated image or noise, which suggests that the visual context might not be exploited by the model at all. We hypothesize that this might be caused by the nature of the commonly used evaluation benchmark, also known as Multi30K, where the translations of image captions were prepared without actually showing the images to human translators. In this paper, we present a qualitative study that examines the role of datasets in stimulating the leverage of visual modality and we propose methods to highlight the importance of visual signals in the datasets which demonstrate improvements in reliance of models on the source images. Our findings suggest the research on effective MMT architectures is currently impaired by the lack of suitable datasets and careful consideration must be taken in creation of future MMT datasets, for which we also provide useful insights.\footnote{Our code and data are available at: \url{https://github.com/jiaodali/vision-matters-when-it-should}.}
\end{abstract}

\section{Introduction}
Multimodal machine translation (MMT) aims to improve machine translation by resolving certain contextual ambiguities with the aid of other modalities such as vision, and have shown promising integration in conventional neural machine translation (NMT) models \cite{specia-etal-2016-shared}. On the other hand, recent studies reported some conflicting results regarding how the additional visual information is exploited by the models for generating higher-quality translations. A number of MMT models \cite{calixto-etal-2017-doubly, helcl-etal-2018-cuni, ive-etal-2019-distilling, lin2020dynamic, yin-etal-2020-novel} have been proposed which showed improvements over text-only models, whereas \newcite{lala-etal-2018-sheffield, barrault-etal-2018-findings, raunak-etal-2019-leveraging} observed that the multimodal integration did not make a big difference quantitatively or qualitatively.
Following experimental work showed that replacing the images in image-caption pairs with incongruent images \cite{elliott-2018-adversarial} or even random noise \cite{wu-etal-2021-good} might still result in similar performance of multimodal models. In light of these results, \citet{wu-etal-2021-good} suggested that gains in quality might merely be due to a regularization effect and the images may not actually be exploited by models during the translation task.

In this paper, we investigate the role of the evaluation benchmark in model performance and whether its tendency to ignore visual information in the input could be a consequence of the nature of the dataset. 
The most widely-used dataset for MMT is Multi30K \cite{elliott-etal-2016-multi30k, elliott-EtAl:2017:WMT, barrault-etal-2018-findings}, which extends the Flickr30K dataset \cite{young-etal-2014-image} to German, French, and Czech translations. 
Captions were translated without access to images, and it is posited that this heavily biases MMT models towards only relying on textual input \cite{elliott-2018-adversarial}. MMT models may well be capable of using visual signals, but will only learn to do so if the visual context provides information beyond the text. For instance, the English word "wall" can be translated into German as either "Wand" (wall inside of a building) or "Mauer" (wall outside of a building), but we find that reference translations in Multi30k are not always congruent with images.

A number of efforts have been put into creating datasets where correct translations are only possible in the presence of images. \newcite{caglayan-etal-2019-probing} degrade the Multi30K dataset to hide away crucial information in the source sentence, including color, head nouns, and suffixes. Similarly, \citet{wu-etal-2021-good} mask high-frequency words in Multi30K. Multisense \cite{gella-etal-2019-cross} collects sentences whose verbs have cross-lingual sense ambiguities. However, due to the high cost of data collection, datasets of such kind are often limited in size. MultiSubs \cite{Wang2021} is another related dataset. which is primarily used for lexical translation because the images are retrieved to align with text fragments rather than whole sentences. 

In this work, we propose two methods to necessitate the visual context --- back-translation from a gender-neutral language (e.g. Turkish) and word dropout in the source sentence. They are simple and cheap to implement, allowing them to be applied on much larger datasets. We test the methods on two MMT architectures and find that they indeed make the model more reliant on the images. 

\section{Method}
In this section, we elaborate two methods to conceal important information in the source textual inputs that can be recovered with the aid of visual inputs. 

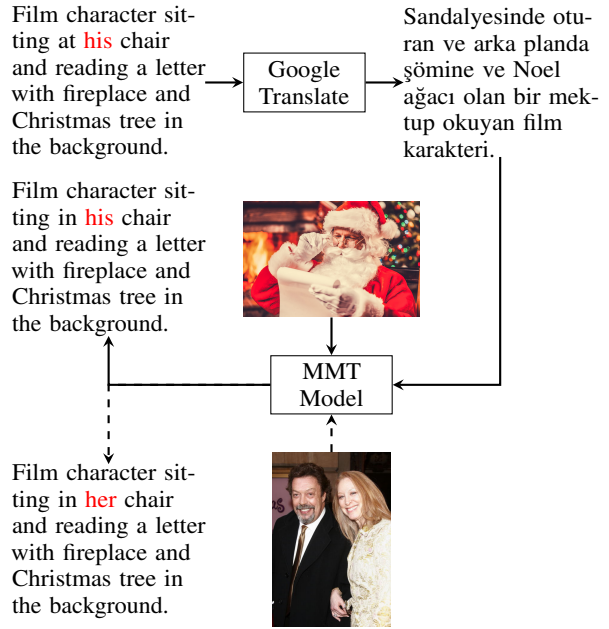
\begin{figure}[t]
\centering\small
\noindent\begin{tikzpicture}[node distance=0.5cm]%
\node[text width=.33\columnwidth, inner sep=0,outer sep=0] (eng) {Film character sitting at \textcolor{red}{his} chair and reading a letter with fireplace and Christmas tree in the background.};
\node[text width=.18\columnwidth, text centered, draw, right=of eng] (nmt) {Google Translate};
\node[text width=.33\columnwidth, right=of nmt, inner sep=0,outer sep=0] (tur) {Sandalyesinde oturan ve arka planda \c{s}\"{o}mine ve Noel a\u{g}acı olan bir mektup okuyan film karakteri.};
\node[text width=.33\columnwidth, below=0.3cm of eng, inner sep=0,outer sep=0] (hyp_true) {Film character sitting in \textcolor{red}{his} chair and reading a letter with fireplace and Christmas tree in the background.};
\node[inner sep=0cm, right=of hyp_true] (img_true) {\includegraphics[width=.3\columnwidth]{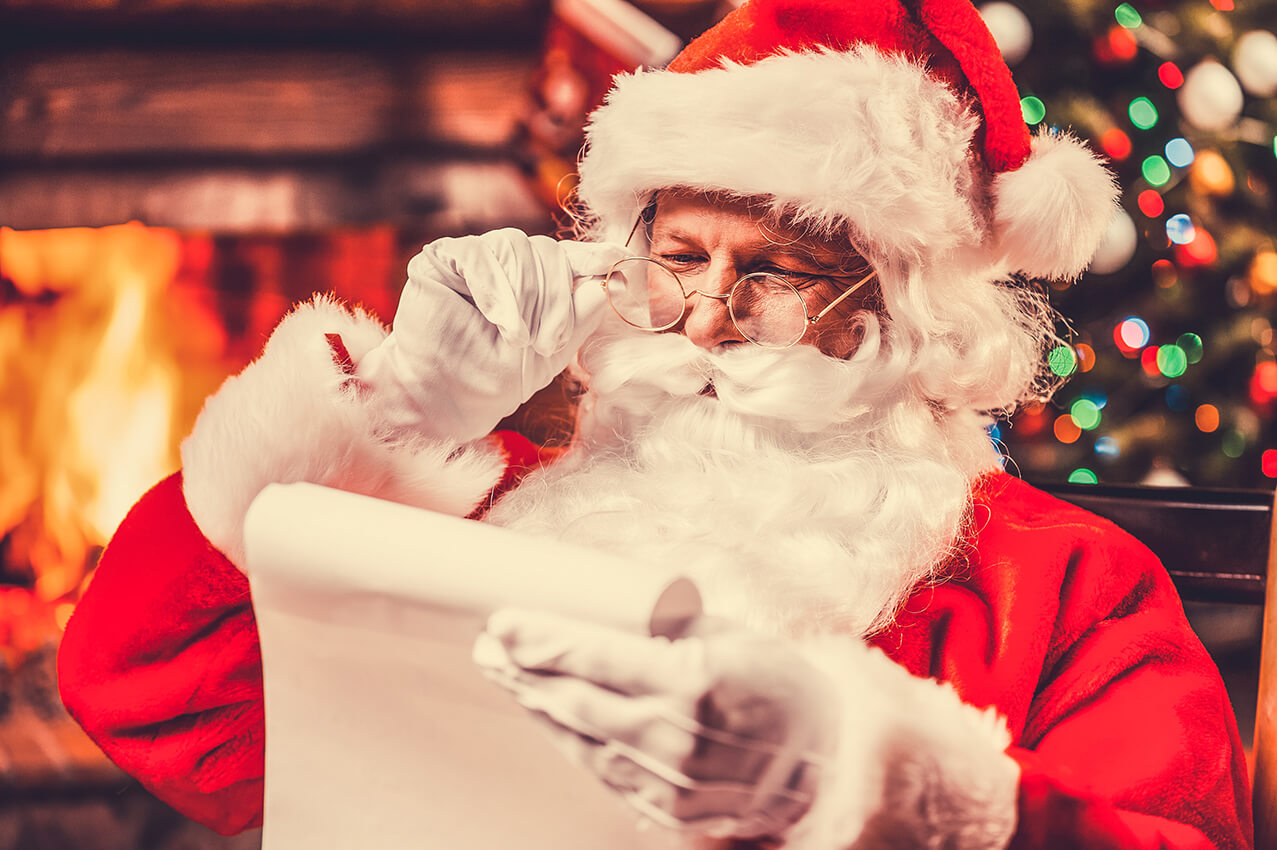}};
\node[text width=.18\columnwidth, text centered, draw, below=of img_true] (mmt) {MMT Model};
\node[inner sep=0cm, below=of mmt] (img_false) {\includegraphics[width=.2\columnwidth]{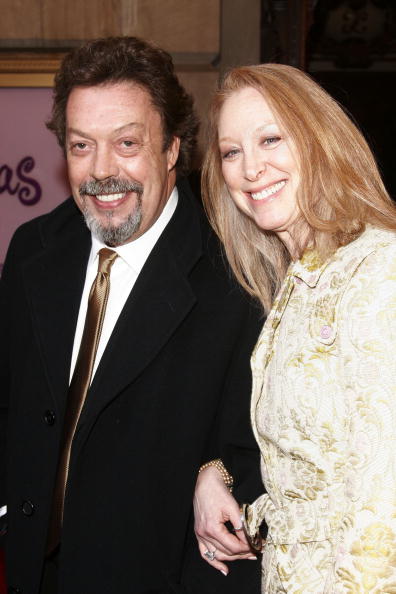}};
\node[text width=.33\columnwidth, below=1.7cm of hyp_true, inner sep=0,outer sep=0] (hyp_false) {Film character sitting in \textcolor{red}{her} chair and reading a letter with fireplace and Christmas tree in the background.};
\draw [arrow] (eng) -- (nmt);
\draw [arrow] (nmt) -- (tur);
\draw [arrow] (tur) |- (mmt);
\draw [arrow] (img_true) -- (mmt);
\draw [arrow] (mmt) -| (hyp_true);
\draw [dashed, arrow] (img_false) -- (mmt);
\draw [dashed, arrow] (mmt) -| (hyp_false);
\end{tikzpicture} 
\caption{An example for back-translation. The image caption is translated into Turkish using a text-only translation system. Then a MMT model is trained to translate it back into English. When an incongruent image is fed into the model, the gender pronoun ``his'' is mistranslated.}
\label{fig:back_translation}
\end{figure}

\paragraph{Back-Translation.} Rather than trying to create reference translations that make use of visual signals for disambiguation, we treat original image captions as the target side and automatically produce ambiguous source sentences.
While such back-translations are generally used for data augmentation \cite{sennrich-etal-2016-improving}, we rely fully on this data for training and testing.
We focus on gender ambiguity, which can be easily created by translating from a language with natural gender (English) into a gender-neutral language (Turkish). In Turkish, there is no distinction between gender pronouns (\textit{e.g.}\ ``he'' and ``she'' are both translated into ``o''). We use a commercial translation system (Google Translate) to translate the image description in English to Turkish. The task is then to translate from Turkish back into English. An example is shown in \cref{fig:back_translation}. 

\paragraph{Word Dropout.} Inspired by \newcite{caglayan-etal-2019-probing}, we degrade the textual inputs to eliminate crucial information. We use a simplified approach that requires no manual annotation, randomly replacing tokens in the source sentence with a special UNK token, subject to a dropout probability $p$ \citep{bowman-etal-2016-generating}.

\section{Experimental Setup}
\subsection{Data Collection}

As our starting point, we use Conceptual Captions \cite{sharma-etal-2018-conceptual}, which contains 3.3M images with captions.
The captions in the dataset have already been processed to replace named entities with hypernyms such as 'person' or profession names such as 'actor'. In order to create a gender-ambiguous dataset we further filter out any sentences containing nouns with information about the gender of the entity (\textit{e.g.}\ woman/man, lady/gentleman, king/queen, etc.) and also remove sentences with professions which are only used in a single gender-specific context (\textit{e.g.}\ `football player', which is always used with the male pronoun in the dataset). We then automatically translate the captions of the resulting dataset into Turkish and use this pseudo-parallel data for training our Turkish-English MMT models. For validation and testing we randomly sample 1000 sentences and use the remaining for training. We refer to this processed dataset as \textbf{Ambiguous Captions} (\textbf{AmbigCaps}). 

For comparison, we also create a Turkish$\rightarrow$English version of Multi30k by back-translating the English side.
\cref{tab:statistics} summarizes the characteristics of the two corpora.

\begin{table}[h]
    \centering
    \small
    \begin{tabular}{c|c|c|c}
        \bf Dataset & \bf \# Sen & \bf \# Words (EN) & \bf \# Gen. PROs\\
        \hline 
        Multi30k  & 31,014 & \phantom{0,}369,048 & \phantom{00}4,181 \\
       \hline
        AmbigCaps & 91,601 & 1,253,400 & 109,440
       
    \end{tabular}
    \caption{Statistical properties (numbers of sentences, words, and gender pronouns in English) of the Multi30k and Ambiguous Captions datasets used in our experiments.}
    \label{tab:statistics}
\end{table}

\subsection{Models}
In our experiments, we consider one NMT model and two MMT models. We follow \newcite{wu-etal-2021-good}'s model and configuration to isolate the cause for the negative results they obtained. We decide not to use the retrieval-based system because it samples images that are not described by the text. We also implement another simple model to demonstrate the applicability of our approaches. 
\paragraph{Transformer.} For text-only baseline, we use a variant of the Transformer that has 4 encoder layers, 4 decoder layers, and 4 attention heads in each layer. The dimensions of input/output layers and inner feed-forward layers are also reduced to 128 and 256 respectively. This configuration has been shown to be effective on Multi30K dataset \cite{wu-etal-2021-good}. The MMT models below follow the same configuration. 

\paragraph{Visual Features.}
Image features are extracted with the code snippet provided by \newcite{elliott-EtAl:2017:WMT},\footnote{\url{https://github.com/multi30k/dataset/blob/master/scripts/feature-extractor}} which uses a ResNet-50 \cite{He:2016} pre-trained on ImageNet \cite{Deng:2009} as image encoder. The `res4\_relu' features $\in\mathbb{R}^{1024\times 14\times 14}$ and average pooled features $\in \mathbb{R}^{2048}$ are extracted. 

\paragraph{Gated Fusion.} Gated fusion model \cite{wu-etal-2021-good} learns a gate vector $\boldsymbol{\lambda}$, and combines textual representation and image representations as follows:
\begin{equation}
    \mathbf{H} = \mathbf{H}_{\text{text}} + \boldsymbol{\lambda} \odot \mathbf{H}_{\text{avg}},
\end{equation}
where $\mathbf{H}_{\text{text}}$ is the output of the Transformer encoder, $\mathbf{H}_{\text{avg}}$ is the average pooled visual features after projection and broadcasting, and $\odot$ denotes the Hadamard product. $\mathbf{H}$ is then fed into the Transformer decoder as in NMT. 

\paragraph{Concatenation.} We implement a different approach to combine textual and visual features. The flattened and projected `res4\_relu' features $\mathbf{H}_{\text{res4\_relu}}$ are directly concatenated with the Transformer encoder representations $\mathbf{H}_{\text{text}}$ as follows:
\begin{equation}
    \mathbf{H} = \left[\mathbf{H}_{\text{text}}; \mathbf{H}_{\text{res4\_relu}} \right].
\end{equation}
This preserves more fine-grained features in the original image and avoids confounding the two modalities.

\subsection{Implementation Details}
We follow \cite{wu-etal-2021-good} and use Adam \cite{Diederik:2015} with $\beta_1=0.9$ and $\beta_2=0.98$. Maximum number of tokens in a mini-batch is 4096. Learning rate warms up from $1e-7$ to $0.005$ in 2000 steps, then decays based on the the inverse square root of the update number. A dropout \cite{Srivastava:2014} of 0.3 and label-smoothing of 0.1 are applied.
The models are trained with early-stopping (patience=10) and the last ten checkpoints are averaged for inference.
We use beam search with beam size 5.
We use the toolkit FAIRSEQ \cite{ott-etal-2019-fairseq} for our implementation.

\subsection{Metrics}
\label{sec:metrics}
\paragraph{BLEU.} We compute the cumulative 4-gram BLEU scores \cite{papineni-etal-2002-bleu} to evaluate the overall quality of translation.  

\paragraph{Gender Accuracy.} Since we are most concerned with the gender ambiguity in the texts, we introduce gender accuracy as an additional metric. We first extract gender pronouns from the sentence. If the sentence contains at least one of the male pronouns $[\text{`he'}, \text{`him'}, \text{`his'}, \text{`himself'}]$, it is classified as `male'; if it contains at least one of the female pronouns $[\text{`she'}, \text{`her'}, \text{`hers'}, \text{`herself'}]$, it is classified as `female'; if it contains both male and female pronouns or neither, it is classified as `undetermined'. We only consider the first two categories,\footnote{See \cref{sec:ethics}.} and compute gender accuracy by comparing the results of references and hypotheses. 

\paragraph{Image Awareness.} To examine models' reliance on the visual modality, we calculate the performance degradation when randomly sampled images are fed. This is also termed as image awareness \cite{elliott-2018-adversarial}.

\section{Results}
The results of our experiments are shown in \cref{tab:results}.

\begin{table*}
\centering\small
\begin{tabularx}{\textwidth}{@{}lYYcYYcYY@{}} 
 \toprule
 \multirow{2}{*}{Model} & \multicolumn{2}{c}{Multi30K (EN$\rightarrow$DE)} & \phantom{} & \multicolumn{2}{c}{Multi30K (TR$\rightarrow$EN)} & \phantom{} & \multicolumn{2}{c}{Ambiguous Captions} \\
 \cmidrule{2-3} \cmidrule{5-6} \cmidrule{8-9}
  & Test2016 \newline BLEU & Multisense \newline BLEU && BLEU & Gender Accuracy && BLEU & Gender Accuracy \\ 
 \midrule
 Transformer & $40.53$ & $26.65$ && $51.64$ & $67.0\%$ && $35.71$ & $73.9\%$ \\
 \midrule
 Gated Fusion & $41.22$\newline$(\uparrow 0.01)$ & $27.09$$\newline$$(\downarrow 0.04)$ && $51.76$\newline$(\uparrow 0.04)$ & $72.2\%$\newline$(\downarrow 0.5\%)$ && $36.68$\newline$(\downarrow 1.71)$ & $80.9\%$\newline$(\downarrow 16.5\%)$\\
 + Word Dropout & $40.65$\newline$(\downarrow 0.19)$ & $26.09$\newline$(\uparrow 0.15)$ && $51.07$\newline$(\uparrow 0.06)$ & $66.1\%$\newline$(\uparrow 0.5\%)$ && $35.35$\newline$(\downarrow 1.28)$ & $79.3\%$\newline$(\downarrow 16.1\%)$\\
 \midrule
 Concatenation & $39.86$\newline$(\uparrow 0.02)$ & $25.71$\newline$(\uparrow 0.25)$ && $51.34$\newline$(\downarrow 0.25)$ & $72.2\%$\newline$(\uparrow 1.4\%)$ && $37.39$\newline$(\downarrow 2.08)$ & $79.4\%$\newline$(\downarrow 18.1\%)$ \\
 + Word Dropout & $40.07$\newline$(\downarrow 0.50)$ & $25.72$\newline$(\downarrow 0.07)$ && $50.81$\newline$(\downarrow 0.90)$ & $68.7\%$\newline$(\downarrow 3.5\%)$ && $35.55$\newline$(\downarrow 2.10)$ & $79.0\%$\newline$(\downarrow 18.2\%)$ \\
 \bottomrule
\end{tabularx}
\caption{Models' performance on various datasets. In the parenthesis is the drop when incongruent images are used (i.e. image awareness). We take the average of 5 runs, each with a different random seed. $\uparrow$ indicates the performance improves after the images are shuffled; $\downarrow$ otherwise.}
\label{tab:results}
\end{table*}

\subsection{Multi30K EN$\rightarrow$DE}
\label{sec:multi30k}
\paragraph{Test2016} We found our MMT models provide little to no improvement over the text-only Transformer. Moreover, the impact of feeding MMT systems with incongruent images is negligible. Our observations conform with previous work \cite{lala-etal-2018-sheffield, barrault-etal-2018-findings, wu-etal-2021-good}, namely that visual signals are not utilized.
\paragraph{Multisense} We also evaluate models trained on Multi30K on the Multisense test set \cite{gella-etal-2019-cross}. Similarly, no substantial difference is observed whether congruent or incongruent images are used. This suggests that it is not just a matter of the Test2016 test set containing too little textual ambiguity, but that the model has not learned to incorporate the visual information necessary for Multisense.\footnote{We also note that some senses in Multisense are rare or unseen in Multi30k.}

\subsection{Multi30K TR$\rightarrow$EN}
Our experiments on the TR$\rightarrow$EN version of Multi30K that we created do not show any substantial improvements in image awareness, which we attribute to the relative sparsity of gender ambiguity at training and test time (see \cref{tab:statistics}).

\subsection{Ambiguous Captions}
Training the same multimodal models on the Ambiguous Captions dataset results in substantial improvements in terms of both BLEU scores and gender accuracy compared to our text-only baseline. 
This suggests that the high level of textual ambiguity in this dataset encourages MMT models to exploit visual information. We further test this hypothesis by repeating the experiment when images are shuffled, and observe that their performance substantially deteriorates, 
especially their ability to infer the correct gender pronouns. For instance, the gated fusion model has an impressive gender accuracy of $80.9\%$ compared to $73.9\%$ of the text-only Transformer, while it drops to $64.4\%$ when incongruent images are used.

We find that both the gated fusion and concatenation model behave similarly, indicating that the choice of dataset has a bigger effect on the success of multimodal modeling than the specific architecture.

\label{sec:conceptual}

\subsection{Effect of Word Dropout}
\label{sec:dropout}
We found word dropout tends to increase image awareness for the concatenation model. This is most evident for Multi30K (TR$\rightarrow$EN), where image awareness increases by $\approx 300\%$. For the gated fusion model, although word dropout leads to more differences in translations between congruent and incongruent image-text alignments (e.g. on Multi30K (TR$\rightarrow$EN), $20$ differences without dropout, $192$ with dropout), it is not well reflected by the image awareness metric. The reason remains to be further inspected. 

Despite having the desired effect of increasing image awareness on the concatenation model, we observe some deterioration of BLEU and gender accuracy compared to the model trained without word dropout; still, we hope that our results serve as a proof-of-concept to motivate future research on regularization schemes that aim to (re)balance visual and textual signal. We note the success of work done in parallel to ours that applied word dropout to increase context usage in context-aware machine translation \citep{fernandes-etal-2021-measuring}.

\section{Conclusion}

Our experiments explain recent failures in MMT, and show that the models we examine successfully learn to rely more on images when textual ambiguity is high (as in our back-translated Turkish--English dataset) or when textual information is dropped out.
Our results suggest that simple MMT models have some capacity to integrate visual and textual information, but their effectiveness is hidden when training on datasets where the visual signal provides little information. 
In the long term, we hope to identify real-world applications where multimodal context naturally provides a strong disambiguation signal.
For the near future, we release our dataset and encourage researchers to utilize it to validate future research on multimodal translation models.
For example, we are interested under which conditions multimodal models learn to exploit visual signal: does the absolute frequency of examples with textual ambiguity matter more, or their proportion?

\section{Broader Impact Statement}
\label{sec:ethics}
Our dataset inherits biases from the Conceptual Captions dataset. We cannot rule out gender bias in the dataset similar to the one described by \newcite{zhao-etal-2017-men}, with males and females showing different distributions, and we only studied a subset of captions with unambiguously male or female pronouns. Despite potential issues with our dataset (which we consider unsuitable for use in production because of aggressive filtering), we believe our work on improving MMT has a positive effect on gender fairness, since multimodal systems with audiovisual clues have the potential to reduce gender bias compared to systems that only rely on textual co-occurrence frequencies. %

\section*{Acknowledgments}
We would like to thank the anonymous reviewers and meta-reviewer for their comments.
This project has received support from the Swiss National Science Foundation (MUTAMUR; no.\ 176727).

\bibliography{anthology,custom}
\bibliographystyle{acl_natbib}

\end{document}